\documentclass[]{article}
\usepackage{proceed2e}

\usepackage{times}

\usepackage[english]{babel}
\usepackage[utf8x]{inputenc}
\usepackage[T1]{fontenc}
\usepackage{enumitem}

\usepackage{natbib}
\usepackage{amsmath}
\usepackage{graphicx}
\usepackage{subcaption}
\usepackage{dsfont}

\usepackage{float}
\usepackage{graphicx}
\usepackage{epstopdf, epsfig}
\usepackage{flushend}
\usepackage[dvipsnames]{xcolor}
\usepackage{amsmath}
\usepackage{latexsym}
\usepackage{caption}
\usepackage{color,soul}
\usepackage{physics}
\usepackage[normalem]{ulem}

 \usepackage{physics}
 \usepackage{amssymb}
 \usepackage{amsmath}

\graphicspath{{figures/}}
\usepackage[colorlinks=true, allcolors=blue]{hyperref}
\usepackage{xcolor}
\usepackage{pgfplots}
\usepackage{bm}
\usepackage{todonotes}
\usepackage{comment}

\usepackage{algorithm,amsmath,algpseudocode,algcompatible,setspace}

\definecolor{light-yellow}{rgb}{1,1,0.75}

\title{Structure and Distribution Metric for Quantifying the Quality of Uncertainty: \\ Assessing Gaussian Processes, Deep Neural Nets, and Deep Neural Operators for Regression}

\author{ {\bf Ethan Pickering} \thanks{Corresponding author: pickering@mit.edu} \hspace{0.5cm} {\bf Themistoklis P. Sapsis} \\
Mechanical Engineering \\ Massachusetts Institute of Technology \\ Cambridge, MA 02139}

\begin{document}

\maketitle

\begin{abstract}
        
        We propose two bounded comparison metrics that may be implemented to arbitrary dimension in regression tasks. One quantifies the \textit{structure} of uncertainty and the other quantifies the \textit{distribution} of uncertainty. The structure metric assesses the similarity in shape and location of uncertainty with the true error, while the distribution metric quantifies the supported magnitudes between the two. We apply these metrics to Gaussian Processes (GPs), Ensemble Deep Neural Nets (DNNs), and Ensemble Deep Neural Operators (DNOs) on high-dimensional and nonlinear test cases. We find that comparing a model's uncertainty estimates with the model's squared error provides a compelling ground truth assessment. We also observe that both DNNs and DNOs, especially when compared to GPs, provide encouraging metric values in high dimensions with either sparse or plentiful data. 

\end{abstract}

\section{Introduction}
Deep neural networks (DNN/NN) have become increasing popular as a surrogate model of choice. This is largely due to their flexibility, propensity for large-data training, and their predictive performance on unseen data. However, the former property, flexibility, comes at a cost. DNNs lack closed analytical forms, rigorous proofs, or unique solutions. Most importantly, there exists no general approach for quantifying the uncertainty (or uncertainty quantification, UQ) for any DNN model. This leads to a hesitancy, reluctance, or a general lack of trust one may have in DNNs, particularly when safety- and mission-critical real world applications are the subject of training and prediction. Finding a general approach for uncertainty quantification for DNNs will greatly remove such concerns and allow modelers to present new data to alleviate large model uncertainties.

However, there are other, and perhaps more fundamental reasons for a lack of DNN uncertainty quantification. What  uncertainty quantification ground truth should a technique be compared to and through what measure should we even compare uncertainty~\citep{gawlikowski2021survey}? Quite often, measures are empirical, qualitative, and low-dimensional. For example, in 1-dimensional problems the ``eyeball'' norm, or intuition, is commonly used to demonstrate the potential of various methods~\citep{lakshminarayanan2016simple,yao2019quality}. This measure is clearly not robust nor scalable to high dimensions, and biased by human perceptions of what is a superior uncertainty. 

Our objective of this study is to propose a predictive-variance and squared-error comparison metric that efficiently quantifies the quality of uncertainty quantification in arbitrarily large dimensions. We propose two scalar metrics for assessing the similarity between uncertainty fields \textit{and} their relationship to the true error. This requires decoupling the structure, where uncertainty and error lie in the parameter space, and distribution of values, i.e. the magnitudes, of uncertainty and error. The former replaces the "eyeball" norm when visualization is impossible, while the latter provides confidence that the magnitudes of predictive variance are reasonable. These metrics allow for us to systemically ask the question: For any given model, what is the model's ``quality of uncertainty''?

This work is also motivated from a Bayesian Experimental Design (BED) and Bayesian Optimization (BO) viewpoint, where measures of the predictive variance are the key ingredient for informing the acquisition of new training data. Thus, the metrics posed aim to answer whether the predictive variances found through various modeling strategies are sufficient for use in BED or BO. Here, we specifically consider Gaussian process (GP) regression, deep neural networks (DNNs/NNs) and deep neural operators (DN0s) for this purpose. Although there is no ``perfect'' quantification of uncertainty, uncertainty quantification found by Gaussian Processes are universally trusted and often cited as the gold-standard and compose the backbone of BED and BO. Despite this, we emphasize that GPs do not generalize well to large-parameter spaces and high dimensions, motivating our curiosity in the structure of uncertainty in DNNs and DNOs.

\section{Uncertainty Metrics}

Several quality of uncertainty metrics exist in the literature, but often these metrics are misleading~\citep{yao2019quality}. Examples include high test log likelihood (LogLL), RMSE, prediction interval coverage probability (PICP)~\citep{pearce2018high}, and mean prediction interval width (MPIW)~\citep{su2018tight}. For high LogLL, the goal is find as much diversity as possible in regions that data has yet to be observed. However, the metric does not possess meaningful bounds and can range from -100 (bad model) to 2 (good model). This is useful for model selection, but not for approximating the true posterior. RMSE only provides performance and greater confidence in the model, not a true uncertainty measure. Similarly, MPIW determines the average width of the 2.5\% to 97.5\% percentile interval with the goal of minimizing MPIW. This metric directly competes with the concept of model diversity and identifying regions of model concern, especially for high-dimensional problems.

Commonly used for ensemble methods, PICP provides a more suitable measure for the structure/quality of uncertainty. PICP directly measures the probability of test data lying within the 2.5\% to 97.5\% percentile interval, where the ideal value is 95\%.  Although PICP provides a probability for capturing test data, it does not provide a measure of how the models do so. PICP does not measure the relative structure of the underlying uncertainty, nor does it weight regions with intriguing uncertainty or test error. As we will show, a large, constant, and ``boring'' predictive variance over a high dimensional space consistently provide large scores, despite providing limited information about the underlying regression problem. The metrics we pose are not fooled by uninteresting uncertainty quantification.




\subsection{Structure Metric: $R$}



The structure metric we implement is no more than the correlation coefficient between the squared error and the predictive variance, but as we will show, it brings far more information than traditionally used metrics, such as PICP or LogLL. Our reasoning for directly comparing squared error and the predictive variance is motivated by BED and BO. For cases where the purpose of uncertainty quantification is to assist in providing models that perform well, in that they generalize to unseen data with small error, we argue the ideal metric be one that measures the ability of uncertainty to identify generalization error. Specifically, our question is whether a model strategy, equipped with a special set of kernels or functions, accurately reflects its perceived errors via its predictive variance. The correlation coefficient does just that, it measures the degree to which the \textit{spatial structure} of the error and the predictive variance are similar.  

To compute the correlation metric, both the squared error, $\epsilon^2(\boldsymbol{\theta})$, where $\boldsymbol{\theta}$ is an $n-$dimensional random variable, and $\sigma^2(\boldsymbol{\theta})$ are evaluated at $\boldsymbol{\theta}_{\mathrm{test}}$ points and represented as a 1-dimensional vector. As is standard for calculating the correlation, we center each vector by its mean and normalize it to unity  such that:
\begin{align}
    {\boldsymbol{\sigma}}^{2T} \boldsymbol{\sigma}^2 &= 1 \nonumber \\
    {\boldsymbol{\epsilon}}^{2T} \boldsymbol{\epsilon}^2 &= 1.
\end{align}
We may then take the inner product of the two normalized vectors to assess their similarity or agreement. Due to the normalization the projection leads to the correlation coefficient, $R$, ranging from -1 to 1,
\begin{equation}
    R = {\boldsymbol{\sigma}^{2T}} \boldsymbol{\epsilon}^2.
\end{equation}

The correlation coefficient brings an unusual set of bounds. A value of 1 indicates that the fields are identical to a scalar multiple, 0 indicates no agreement, or orthogonality between the two methods, while -1 presents vectors that are identically inverse. For comparing $\sigma^2$ and $\epsilon^2$, all 0 and negative values are effectively useless. Negative values are extremely rare and problematic when observed, as we are comparing positive valued fields anchored by identical training datasets. However, the ability of the correlation coefficient to significantly \textit{penalize} inverse behavior between the variance and error is the \textit{attractive feature}. This is specifically what keeps this measure from being tricked by large, constant, and boring predictive variances found by GPs later. As a consequence of the penalization, values should be interpreted as reporting that \textbf{at least} an $R$ fraction of the predictive variance regions/mass is in agreement with the squared error. 

These bounded values provide a clear and interpretable metric that is defined only by the number of query points in an arbitrarily large parameter space (i.e. $\mathbf{\theta}$) and is not hampered by large dimensions. 

\subsection{Distribution Metric: NDIP}

With structure considered, we are also interested in the similarity of supported magnitudes of the predictive variance and pose a metric for assessing the quality of the distribution of uncertainty. Although structure is critical for BED or BO, an understanding of the amplitude of the predictive variance is important for model confidence. In order to define the distribution metric, we remove the notion of structure and look solely at the distribution of predictive variances and squared errors. 

Again, we begin with a test vector of predictive variances and squared errors, but instead of centering and normalizing, we fit the values, using a kernel density estimator, to a model agnostic distribution $p_{\sigma^2}(\sigma^2)$. We then discretize and normalize this distribution such that
\begin{align}
    {\boldsymbol{p}_{\sigma^2}}^{T} {\boldsymbol{p}_{\sigma^2}} &= 1 \nonumber \\
    {\boldsymbol{p}_{\epsilon^2}}^{T} {\boldsymbol{p}_{\epsilon^2}} &= 1.
\end{align}
Taking the inner product of this normalized distribution gives the Normalized Distribution Inner Product (NDIP) 
\begin{align}
    \mathrm{NDIP} = {\boldsymbol{p}_{\sigma^2}}^{T} {\boldsymbol{p}_{\epsilon^2}}.
\end{align}
Just as the correlation coefficient, a value of 1 presents two identical distributions, while a value of 0 gives orthogonality (as all value are positive for probability distributions, negative values are not possible).

\subsection{Surrogate Models} \label{sec:models}
\subsubsection{Gaussian Process Regression} \label{sec:methods_GP}

For low-dimensional problems, Gaussian process (GP) regression \citep{rasmussen2003gaussian} is seen as the ``gold standard'' for Bayesian design and uncertainty quantification. A Gaussian process $\bar{f}(\boldsymbol{\theta})$, where $\boldsymbol{\theta}$ is a random variable, is completely specified by its mean function $m(\mathbf{\boldsymbol{\theta}})$ and covariance function $k\left(\mathbf{\boldsymbol{\theta}}, \mathbf{\boldsymbol{\theta}}^{\prime}\right)$. For a dataset $\mathcal{D}$ of input-output pairs ($\{\mathbf{\boldsymbol{\Theta}}, \mathbf{y}\})$ and a Gaussian process with constant mean $m_{\boldsymbol{0}}$, the random process $\bar{f}(\mathbf{\boldsymbol{\theta}})$ conditioned on $\mathcal{D}$ follows a normal distribution with posterior mean and variance
\begin{equation}
\mu(\mathbf{\boldsymbol{\theta}})=m_{0}+k(\mathbf{\boldsymbol{\theta}}, \mathbf{\boldsymbol{\Theta}}) \mathbf{K}^{-1}\left(\mathbf{y}-m_{0}\right)
\label{eqn:mean_GP}
\end{equation}
\begin{equation}
\sigma^{2}(\mathbf{\boldsymbol{\theta}})=k(\mathbf{\boldsymbol{\theta}}, \mathbf{\boldsymbol{\theta}})-k(\mathbf{\boldsymbol{\theta}}, \mathbf{\boldsymbol{\Theta}}) \mathbf{K}^{-1} k(\mathbf{\boldsymbol{\Theta}}, \mathbf{\boldsymbol{\theta}})
\label{eqn:cov_GP}
\end{equation}
respectively, where $\mathbf{K}= k (\mathbf{\boldsymbol{\Theta}}, \mathbf{\boldsymbol{\Theta}})+\sigma_{\epsilon}^{2} \mathbf{I}$. Equation \eqref{eqn:mean_GP} can be used to predict the value of the surrogate model at any point $\mathbf{\boldsymbol{\theta}}$, and \eqref{eqn:cov_GP} to quantify uncertainty in prediction at that point \citep{rasmussen2003gaussian}. Here, the kernel is chosen as a radial-basis-function (RBF) kernel with automatic relevance determination (ARD),
\begin{equation}
k\left(\mathbf{\boldsymbol{\theta}}, \mathbf{\boldsymbol{\theta}}^{\prime}\right)=\sigma_{f}^{2} \exp \left[-\left(\mathbf{\boldsymbol{\theta}}-\mathbf{\boldsymbol{\theta}}^{\prime}\right)^{\top} \mathbf{L}^{-1}\left(\mathbf{\boldsymbol{\theta}}-\mathbf{\boldsymbol{\theta}}^{\prime}\right) / 2\right],
\end{equation}
where $\boldsymbol{L}$ is a diagonal matrix containing the lengthscales for each dimension and the GP hyperparameters appearing in the covariance function $(\sigma_{f}^{2}$ and $\boldsymbol{L}$ in \eqref{eqn:cov_GP} are trained by maximum likelihood estimation).

\subsubsection{Deep Neural Networks and Operators} \label{sec:methods_NN}

Here we implement the architecture proposed by \cite{lu2021learning} for approximating nonlinear operators: \textit{DeepONet}, only covering the basic details here. DeepONet seeks approximations of nonlinear operators by constructing two deep neural networks, one representing the input function at a fixed number of sensors and another for encoding the ``locations'' of evaluation of the output function. The first neural network, termed the ``branch'', takes input functions, $u$, observed at discrete sensors, $x_i, i = 1 ... m$. The second neural network, the ``trunk'', encodes inherent qualities of the operator, denoted as $z$. Together, these networks seek to approximate the nonlinear operation upon $u$ and $z$ as $G(u)(z) = y $, where $y$ denotes the scalar output from the $u,z$ input pair. Therefore, our set of input-output pairs when discussing DeepONet are, $\{[\mathbf{u},\mathbf{z}], G(\mathbf{u})(\mathbf{z})\}$.

Although NNs are attractive for approximating nonlinear regression tasks, their complexity rids them of analytical expressions for uncertainty. There are several techniques for quantifying uncertainty in neural networks, however we only consider ensemble methods here (see \cite{gawlikowski2021survey} and \cite{psaros2022uncertainty} for comprehensive reviews of methods for uncertainty quantification in NNs). 

Ensemble approaches have been used quite extensively throughout the literature~\citep{hansen1990neural,lakshminarayanan2016simple} and despite their improved results for identifying the underlying tasks at hand~\citep{gustafsson2020evaluating}, their utility for quantifying uncertainty in a model remains a topic of debate. The are several approaches for creating ensembles. These include random weight initialization~\citep{lakshminarayanan2016simple},  different network architectures (including activation functions), data shuffling, data augmentation, bagging, bootstrapping, and snapshot ensembles~\citep{loshchilov2016sgdr,huang2017snapshot,smith2015no} among others. Here we employ random weight initialization, as it has been found to perform similarly or better as BNN approaches (Monte Carlo Dropout and Probablistic Backpropagation) for evaluation accuracy and out-of-distribution detection for both classification and regression tasks~\citep{lakshminarayanan2016simple}. Although much of the literature is skeptical of the generality of ensembles to provide uncertainty estimates, recent viewpoints, specifically \cite{wilson2020bayesian}, have argued that DNN ensembles provide a very good approximation of the posterior. 

We train $N=10$ randomly weight-initialized NN models, each denoted as $\tilde{G}_{n}$, that find the associated solution field $Y$ for inputs $u$ and $z$. This allows us to then determine the point-wise variance of the models as
\begin{equation}
    \sigma^2(u,z) = \frac{1}{(N-1)} \sum_{n=1}^{N} (\tilde{G}_{n}(u)(z) - \overline{\tilde{G}(u)(z)})^2 
\end{equation}
where $\overline{\tilde{G}(u)(z)}$ is the mean solution of the model ensemble and $N$ are the total number of models retained from the initialized weight models. Finally, we must adjust the above representation to match the description for GPs and to permit a systematic scaling in dimension. The input parameters, $\boldsymbol{\theta}$, represent the union of two set of parameters, the stochastic parameters $\boldsymbol{\theta}_u$ and the operation parameters $\boldsymbol{\theta}_z$. The parameters $\boldsymbol{\theta}_u$ typically represent coefficients to a set of functions that represent a decomposition of a random function $u = \boldsymbol{\theta}_u \mathbf{\Phi}(x_1, ... x_m)$, while $\boldsymbol{\theta}_z = z$ represent non-functional parameters. Thus, the DNN/DNO description for UQ may be recast as:
\begin{align}
    \sigma^2(\boldsymbol{\theta}) &= \sigma(\boldsymbol{\theta}_u \cup \boldsymbol{\theta}_z) \\ \nonumber &=  \frac{1}{(N-1)} \sum_{n=1}^{N} (\tilde{G}_{n}(\boldsymbol{\theta}_u \mathbf{\Phi}(x_1, ... x_m))(\boldsymbol{\theta}_z) \\ \nonumber &- \overline{\tilde{G}(\boldsymbol{\theta}_u \mathbf{\Phi}(x_1, ... x_m))(\boldsymbol{\theta}_z)})^2.
\end{align}
For this study, a modest 10 randomly initialized ensemble members are used. 

\section{Results}
Here we demonstrate the metric on two applications, a stochastic oscillator (SO) \citep{mohamad2018sequential,blanchard2020output} and a version of the nonlinear Schr\"{o}dinger equation (NLS) \citep{majda1997one}. Details for each set of equations and the appropriate output definitions are given in Appendix~\ref{app:Applications}.

We are specifically interested in how uncertainty is quantified for low- to high-dimensional stochastic processes and with regard to sparsely and densely populated training sets. The dimension of the stochastic excitation (SO, $u(t)$) or initial condition (NLS, $u(x)$),  is defined by a finite number of random variables using the Karhunen-Loève expansion,
\begin{equation}
    u \approx \boldsymbol{\theta}_u \mathbf{\Phi},
\end{equation}
where $\boldsymbol{\theta}_u \in \mathbb{R}^{m}$ is a vector of random variables and  $\mathbf{\Phi}$ are the eigenvectors of an associated correlation matrix. This definition allows a systematic increase in the input space, on a [-6,6] domain, to arbitrarily large dimensions.

\begin{figure}
    \centering
    Oscillator ($2D$) \hspace{1.5cm}  NLS ($2D$) \phantom{PP}
\includegraphics[width=0.475\textwidth,trim={0cm 0cm 0cm 0cm},clip]{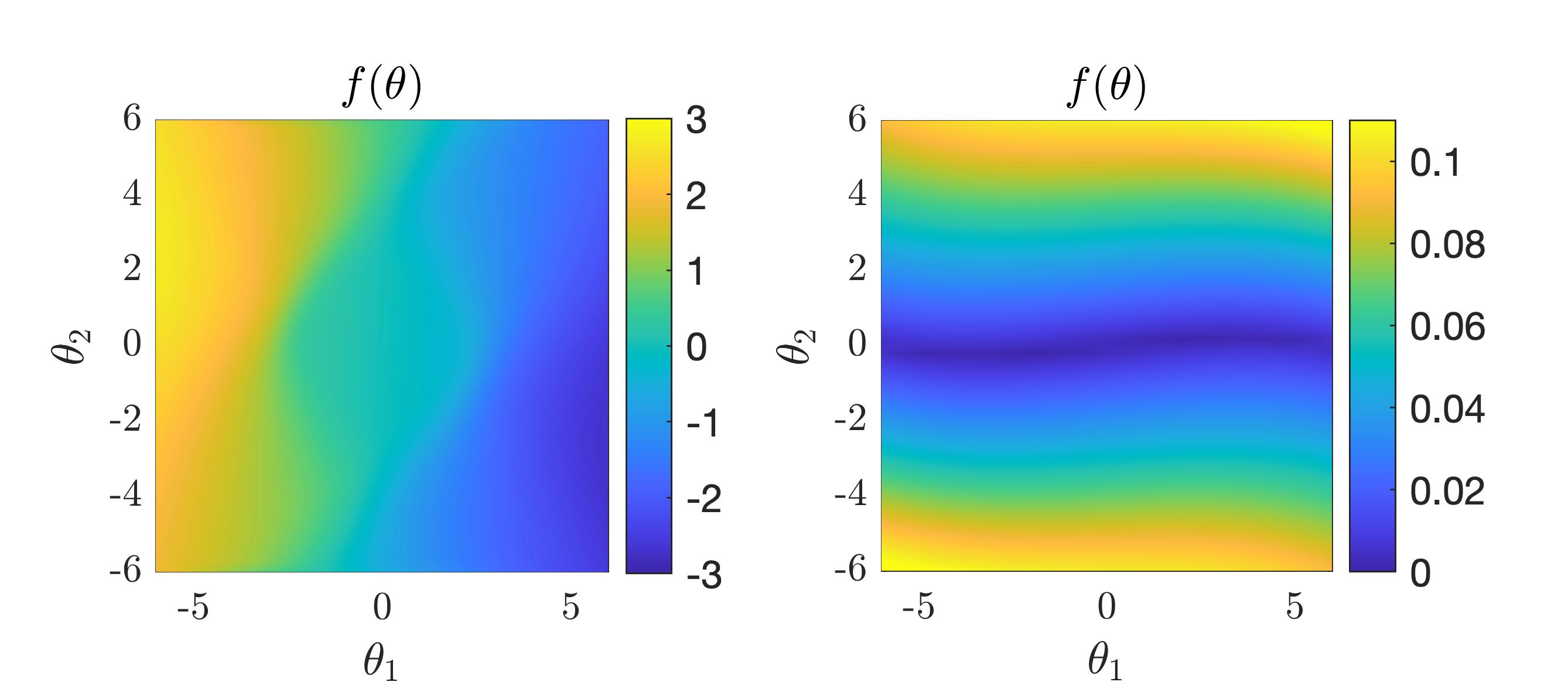}
\caption{The objective functions, as shown by the contour values, for the 2-dimensional stochastic forcing applied to the SO (left) and the stochastic initial condition to the NLS (right).}
    \label{fig:Stoch_Objective}
\end{figure}

\begin{figure}
    \centering
    $a)$ \textit{Stochastic Oscillator} \\ \vspace{0.25cm}
Gaussian Process
\includegraphics[width=0.5\textwidth,trim={4cm 45cm 0cm 0cm},clip]{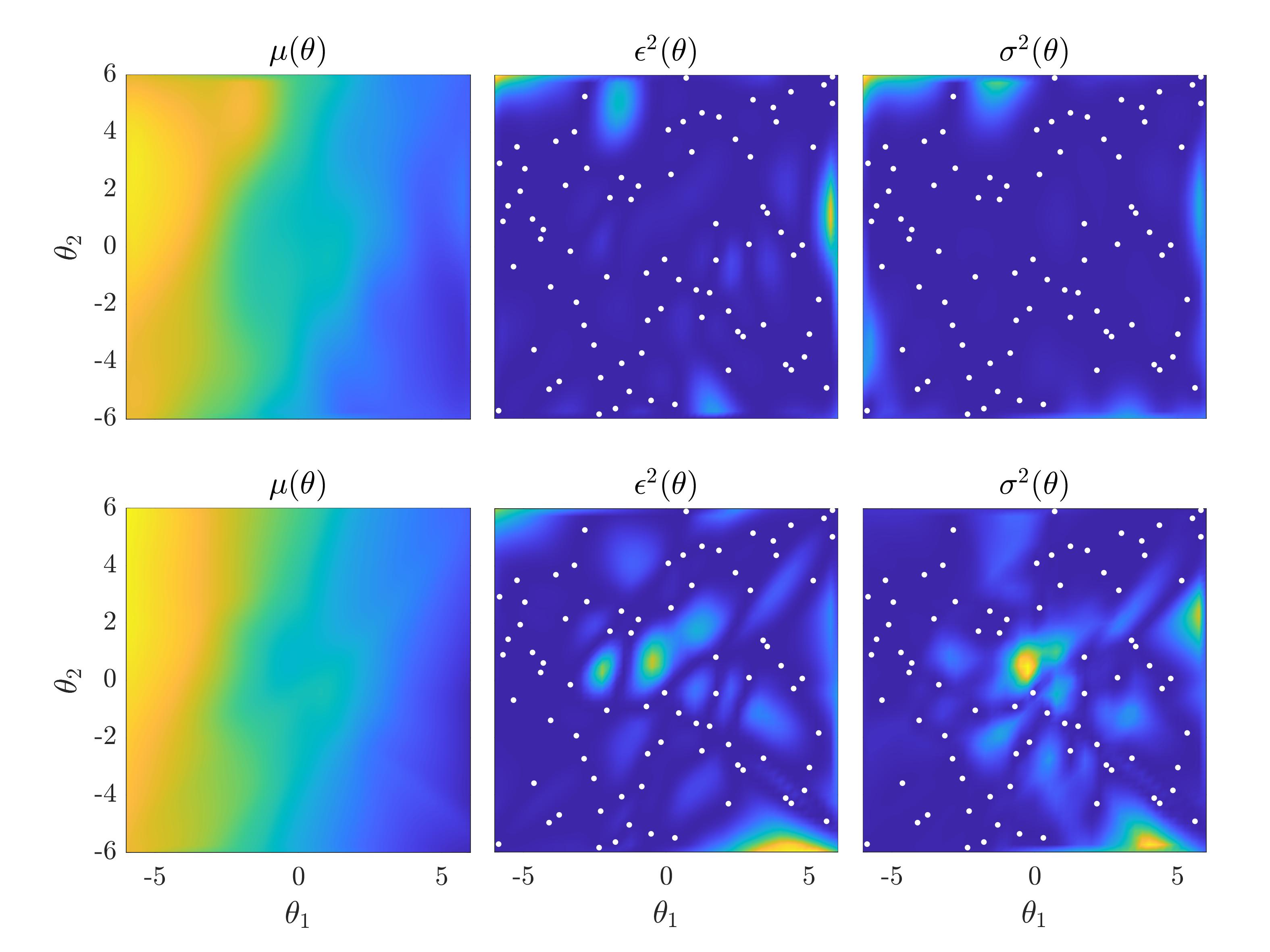}
Deep Neural Network
\includegraphics[width=0.5\textwidth,trim={4cm 0cm 0cm 40cm},clip]{figs/GP_NN_Oscil_2D_Landscape.jpg} \\
$b)$ \textit{Nonlinear Schr\"{o}dinger Equation} \\  \vspace{0.25cm}
Gaussian Process
\includegraphics[width=0.5\textwidth,trim={4cm 45cm 0cm 0cm},clip]{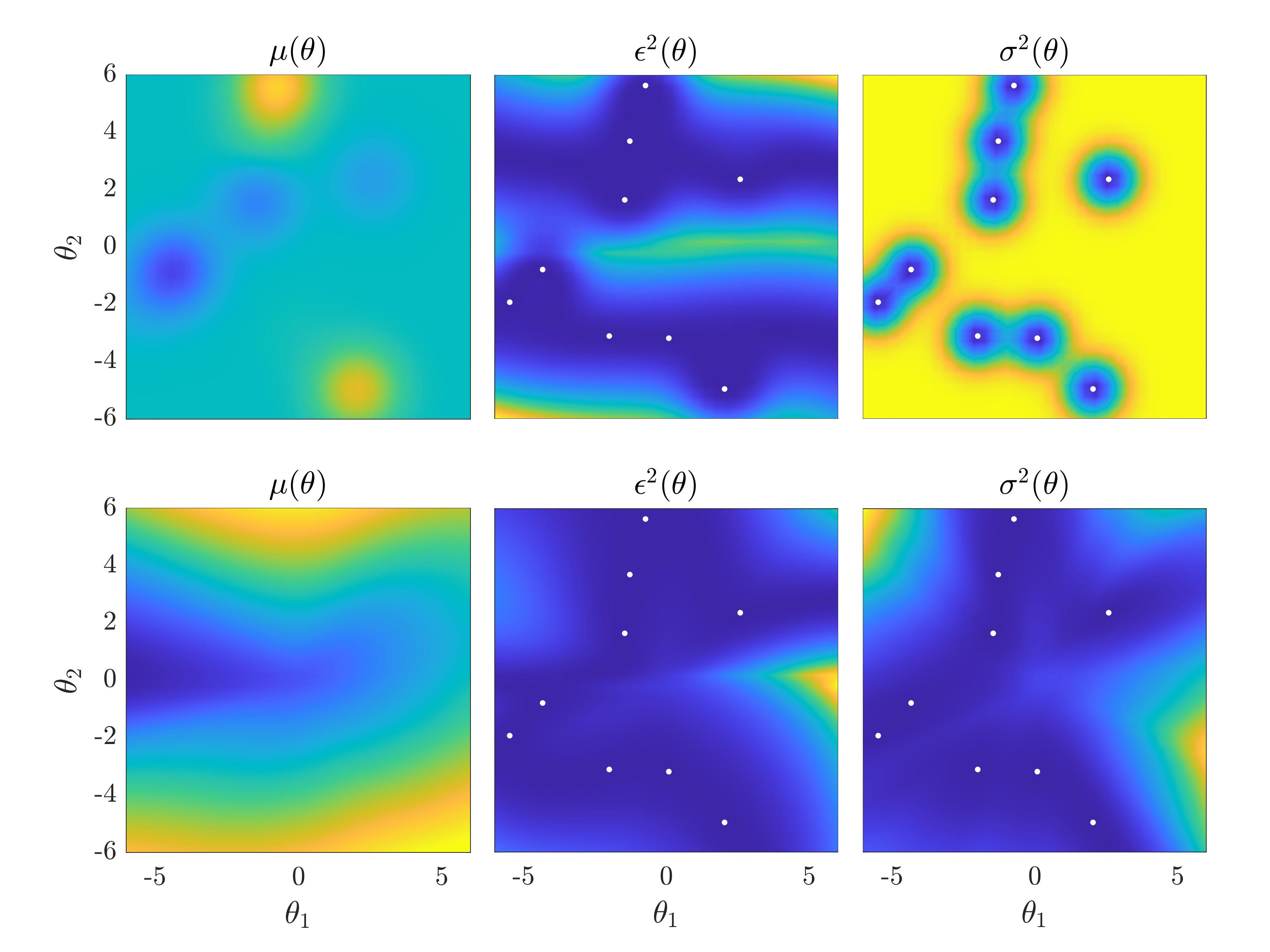}
Deep Neural Network
\includegraphics[width=0.5\textwidth,trim={4cm 0cm 0cm 40cm},clip]{figs/GP_NN_MMT_2D_Landscape_9.jpg}
\caption{$2D$ GP and NN regressions, squared errors, and predictive variances for SO $a)$ with dense training data ($\boldsymbol{\theta}_n=81$) and NLS $b)$ with sparse training data ($\boldsymbol{\theta}_n=9$).}
    \label{fig:2D_Oscil}
\end{figure}

\begin{table}
\caption{Metric values for the visualized squared errors and predictive variances in figure \ref{fig:2D_Oscil}. ``Better'' scores are bold for each case.}
\label{tab:Vals}
\small{
\begin{tabular}{c||c|c|c|c|c|c}
    \textbf{SO}&  $R$ & NDIP & PICP & LogLL & MIWP & MSE \\
    \hline
    GP & \textbf{0.76} & \textbf{0.99} & \textbf{0.55} & \textbf{-1.65} & 0.12 & 0.009 \\
    NN & \textbf{0.76} & 0.98 & 0.46 & -3.24 & \textbf{0.07} & \textbf{0.004} \\
   \textbf{ NLS} \\
    \hline
     GP & 0.45 & 0.42 & \textbf{0.88} & \textbf{2.31} & 0.046 & 0.19 \\
     NN & \textbf{0.69} & \textbf{0.93} & 0.14 & -1.09 & \textbf{0.007} &\textbf{ 0.06}
\end{tabular}
}
\end{table}

\subsection{Gaussian Processes and Deep Neural Networks}
\subsection{$2D$ Example}

For the 2D case, we provide visual examples of our approach. Figure \ref{fig:Stoch_Objective} presents the objective output, or regression task, for the 2-dimensional stochastic representations for both the SO and NLS. Considering that only stochastic dimensions, $\boldsymbol{\theta}_u$, are considered, the DeepONet architecture reduces to a standard DNN (or simply NN).

Figure \ref{fig:2D_Oscil}~$a)$ provides the estimated maps, the squared error, and the predictive uncertainties for SO (81 training points), NLS (9 training points) using both GPs and DNNs, while table \ref{tab:Vals} provides the scores from our two metrics and the four metrics featured in \cite{yao2019quality} (defined in Appendix~\ref{app:Metrics}). Immediately from the figure and the reported $R$ value, we observe that both the GP and NN provide predictive variances that agree similarly in structure with the squared error. The similarity is so close that their $R$ value is 0.76 for both models, despite significant differences between the GP and NN predictive variances. Even with the visual and $R$ value agreement, the traditionally trusted PICP and LogLL favor the GP model, while the PICP for both GP and NN is quite low, and would be considered unacceptable. For this case, PICP fails to capture the relationship between true error and predictive variance.

The visualization and metrics of the sparsely trained NLS example tell an alternate story with a similar conclusion: The standard metrics cannot identify agreement in structure between the predictive variance and squared error. Here the $R$ value for the NN is 0.69 and visual inspection confirms a similar structure, but the PICP is only 0.14 and the LogLL underwhelms the GP case by a factor of $e^3$. The GP, however, possess a reasonable $R$ at 0.45, but a substantially large PICP of 0.88. Clearly, the GP model provides a predictive uncertainty that is less informative to the underlying error than the NN. PICP is susceptible to constant uncertainty. This will be exacerbated in higher dimensions.

\begin{figure}
    \centering
$a)$ \textit{Stochastic Oscillator} \\ \vspace{0.25cm}
\includegraphics[width=0.45\textwidth, trim={3.5cm 0cm 3cm 0cm},clip]{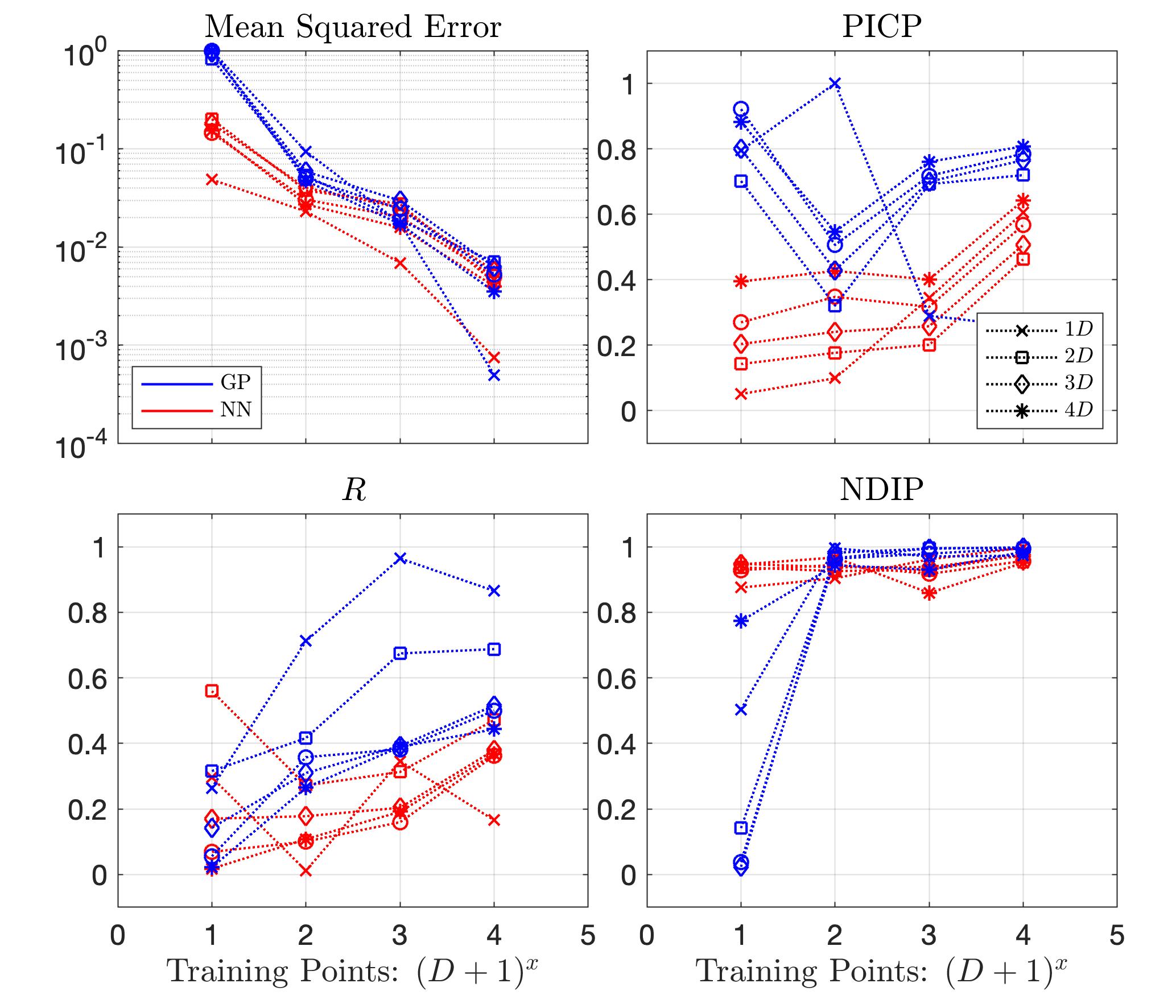} \\
\vspace{0.25cm}
$b)$ \textit{Nonlinear Schr\"{o}dinger Equation} \\  \vspace{0.25cm}
\includegraphics[width=0.475\textwidth, trim={3.5cm 0cm 3cm 0cm},clip]{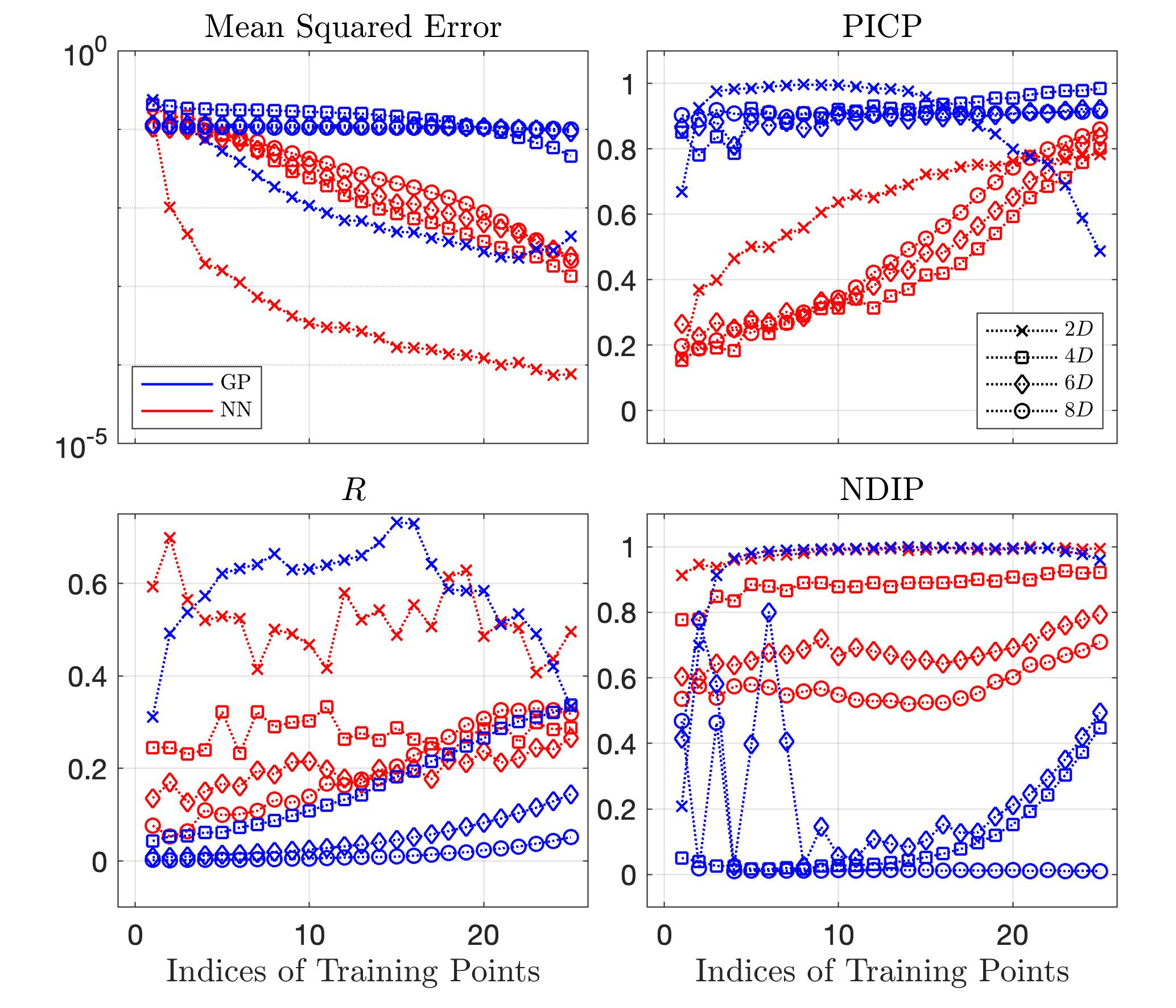}
    \caption{Median values of the MSE, PICP, $R$ and NDIP from sparse to dense training points for many dimensions. $a)$ the stochastic oscillator (25 experiments) and $b)$ the Nonlinear Schr\"{o}dinger equation (10 experiments).}
    \label{fig:MMT_Oscil_Projs}
\end{figure}

\subsection{Scaling: Dimensions and Data}

We now look to explore the metrics as we increase dimension and vary the training data from sparse to dense. Here we only compare MSE, PCIP, $R$, and NDIP, as LogLL and MPIW provided consistently monotonic values with increasing data (LogLL increasing, MPIW decreasing), and limited insight for varied dimensionality. 

For the SO case, we provide the median results of 25 independent experiments from 1-5$D$ and training points ranging from $n_{\mathrm{train}} = (D+1)^x$ where $D$ is the dimension and $x = [1,2,3,4]$ with $x=1$ relating to a sparse distribution and $x=4$ being a dense distribution (e.g. $D=2$ gives $n_{\mathrm{train}} = [3,9,27,81]$.). The metrics are evaluated over $10^3$ test points for $1D$, $10^4$ for $2D$ and $10^5$ for $3-5D$. All training and test points are computed using Latin hypercube sampling (LHS). Figure \ref{fig:MMT_Oscil_Projs}~$a)$ presents the median metrics related to both models (GP as blue and NN as red) and all five dimensions. We can see that the $R$ values agree with the visual evidence presented earlier and PICP tracks similarly to $R$ for this example. Juxtaposing all subplots it is not clear which modeling approach is ``better''. NNs provide better MSE, GPs provide slightly better structural metric values over all cases (especially low-$D$), and NNs perform better in both metrics for sparse data. Based on these results, if one is happy with one model's uncertainty quantification, there is no justifiable reason to not be just as happy with the other.

Turning to the more complex NLS example in \ref{fig:MMT_Oscil_Projs}~$b)$, we test 10 independent experiments over 25 log-spaced training points, $2D$: 3-300, $4D$:5-1000, $6D$:7-2500, $8D$:9-5000, where the x-axis denotes the indice of the interval. The MSE shows that for $4D$ and higher, the GPs are breaking down due to the complexity of the map and high-dimensional regression. Despite this, the PICP for GPs at all dimensions and training sizes, except $2D$ at large $n_{\mathrm{train}}$, is nearly 1. The $R$ metric is not fooled by the poor GP regression and gives values of nearly zero for $>2D$, until approximately $>10^3$ points are provided. This is a significant amount of training data for GPs. Additionally, the NDIP metric begins to take action. The low NDIP scores for GPs at $>2D$ stress that the predictive variance does not present the rich distribution of underlying error values. The NNs, however, provide relatively impressive scores for $R$, NDIP, and MSE, for all dimensions. Only PICP reports poor NN uncertainty quantification.

\subsection{Deep Neural Operator: DeepONet}
\begin{figure}[!]
    \centering
    NLS Realization
\includegraphics[width=0.475\textwidth, trim={3cm 0cm 2cm 0cm},clip]{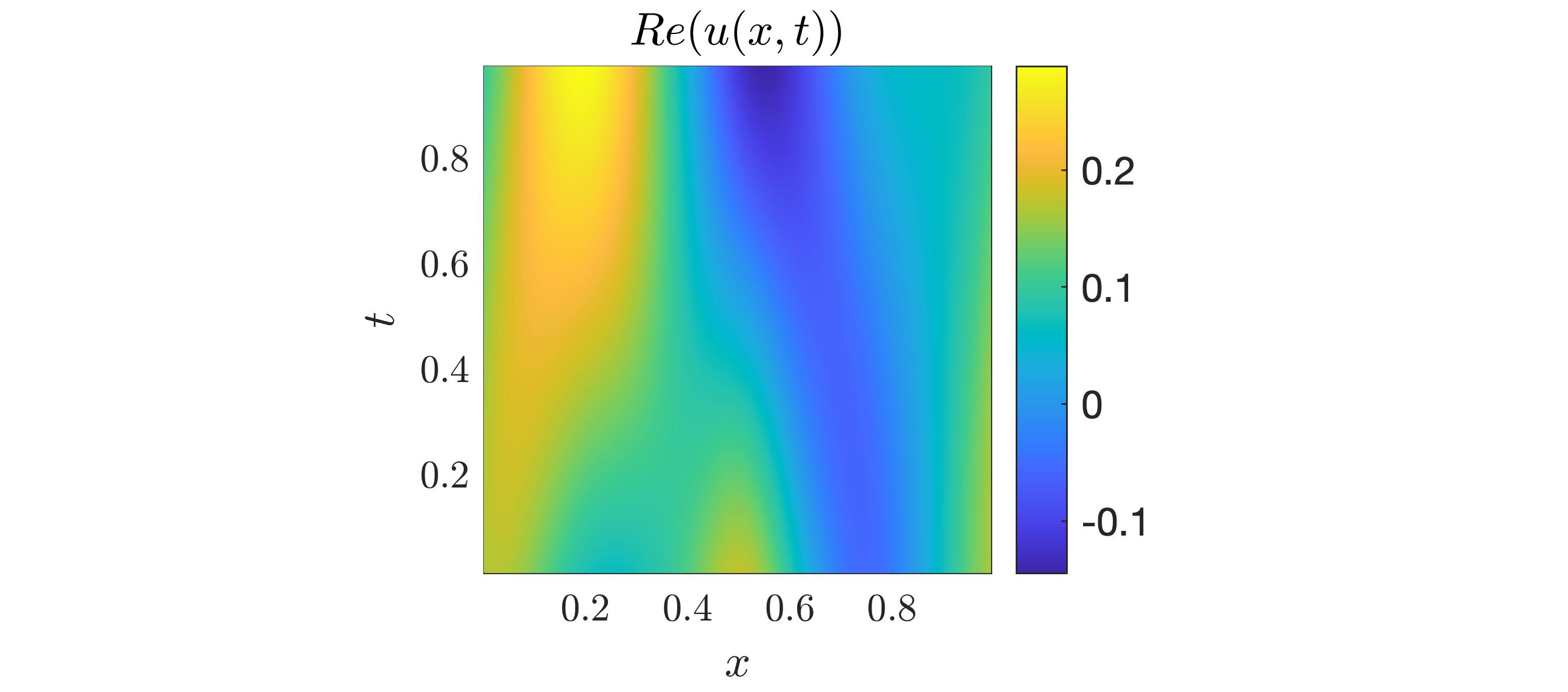}
\caption{One realization (i.e. $\boldsymbol{\theta}$) of the NLS solution.}
    \label{fig:MMT_Op_Realization}
\end{figure}

\begin{figure}[!]
    \centering
    \textit{Deep Neural Operator} \\
    $a)$ In Realization Training Points: 10\\
\includegraphics[width=0.475\textwidth, trim={0cm 0cm 0cm 0cm},clip]{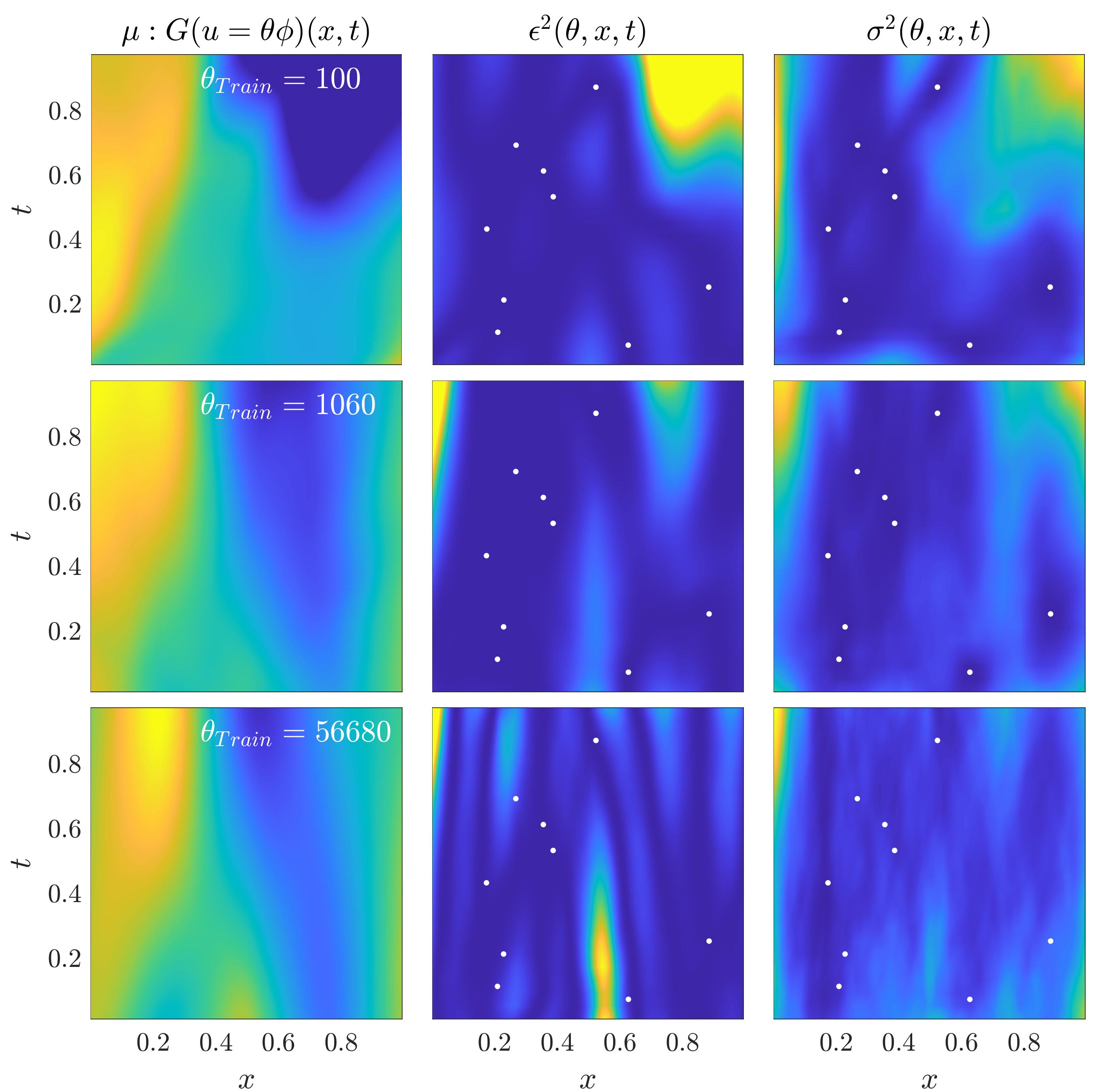} \\ 
$b)$ In Realization Training Points: 0\\
\includegraphics[width=0.475\textwidth, trim={0cm 0cm 0cm 0cm},clip]{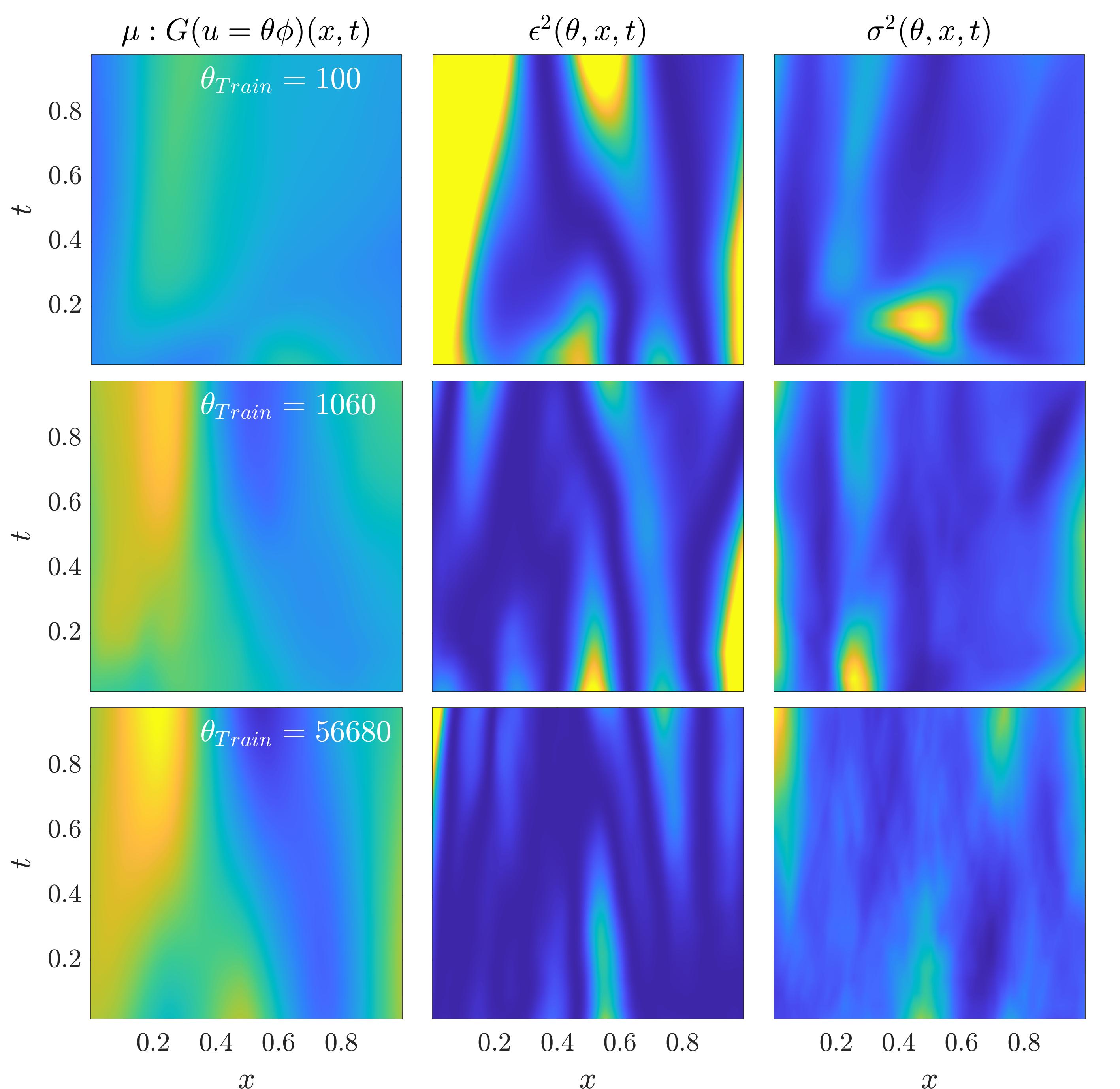}
\caption{Mean regression, squared error, and predictive variance over $x,t$ for one realization of $\boldsymbol{\theta}$ with 100, 1060, and 56680 training samples, top to bottom, respectively. $a)$ provides training that included 10 points from the visualized realization, while $b)$ does not. $R$ from top to bottom: $a)$ 0.72, 0.61, 0.42; $b)$ -0.18, 0.24, 0.57.}
    \label{fig:MMT_Op_Realization_Train}
\end{figure}

\begin{figure}[!]
    \centering
\includegraphics[width=0.46\textwidth, trim={3.5cm 0cm 3cm 0cm},clip]{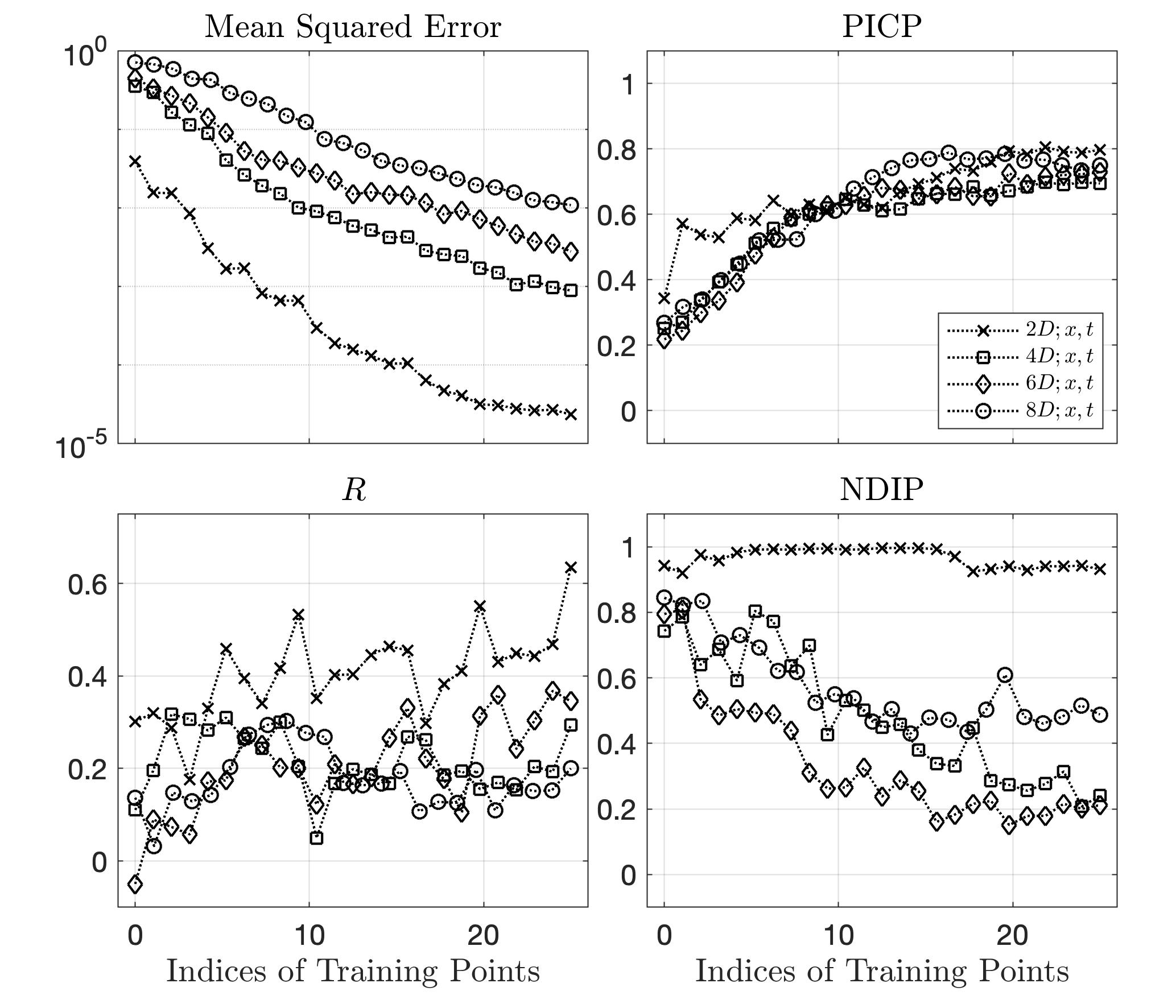}
\caption{Median values of the MSE, PICP, $R$ and NDIP from sparse to dense training points for the operator case.}
    \label{fig:MMT_Op_Projs}
\end{figure}


We now extend our analysis for deep neural operators for the NLS case. Instead of only regressing on a set of random initial conditions from $2-8D$, we add both the spatial and temporal dimensions, $x,t$. This is a particularly challenging regression. Considering the difficulty of GPs to parameterize the non-operator case, without $x,t$, we only consider the uncertainty of an ensemble of DeepONets.

Figure~\ref{fig:MMT_Op_Realization} visualizes one realization of an $8D$ initial condition propagated in time and space, while \ref{fig:MMT_Op_Realization_Train} provides two sets of training snapshots for the same realization. Figure \ref{fig:MMT_Op_Realization_Train}~$a)$ provides the mean DNO regressions when 10 randomly chosen points from this realization are include in the training set, denoted as white points. Regressions are given for $\theta_{\mathrm{train}} = [100, 1060, 56650]$, where $\theta_{u,\mathrm{train}} =\theta_{\mathrm{train}}/10$ as each function set of parameters includes 10 randomly chosen $x,t$ points. We observe that with little training data, the squared error and predictive variance are small around the training points, but as training increases the individual influence of these points diminishes. Interestingly, similar errors and variances appear for \ref{fig:MMT_Op_Realization_Train}~$b)$ when training increases, despite the absence of training points from this realization.

With a visualization of the task at hand, we apply the metrics over the various dimensions and with increasing data size. Here the training data sizes are log-spaced over an interval with 25 indices, $2D;x,t$: 50-10000, $4D;x,t$:70-25000, $6D;x,t$:90-50000, $8D;x,t$:110-75000. Figure \ref{fig:MMT_Op_Projs} provides the median metric values for 10 independent experiments. Despite the dimensionality and the large data sizes considered, both the $R$ and NDIP values are promising. For the $R$ value it is particularly intriguing that large data sizes produce substantial agreement with the underlying error. This implies that even at these scales, BED and BO with DNOs is likely fruitful. The NDIP metric does show reduction with increased data size at large dimension, however, the values appear to converge. This suggests the variance and error distributions are not substantially disconnected. Although the PCIP value increases monotonically with data size, our previous observations leave little room for prescribing meaning to these trends. 


\section{Conclusions}
\textbf{The proposed correlation metric computed over predictive uncertainty and squared error provides a representative measure a model's underlying predictive deficiencies.}
The metric consistently quantifies the similarity in structure between uncertainty and error, while other metrics do not. Metrics, such as PICP, appear to provide useful information at low dimensions, but when complexity and dimensionality are increased these metrics are easily fooled (e.g. figure \ref{fig:MMT_Oscil_Projs}~$b)$). The correlation metric provides a means for quantifying the relationship of the topology between predictive variance and error. Such a measure is critical for instilling confidence that the surrogate model of choice is performing well. This is especially important for Bayesian experimental design and optimization that rely on intriguing predictive variance estimates for efficient data acquisition.  

\textbf{Comparing the uncertainties emitted by GPs, DNNs, and DNOs, or other models, unjustly biases what a ``good'' uncertainty is for a given model.} Often, models such as GPs or Hamiltonian Monte Carlo \citep{neal2011mcmc} are lauded for ideal uncertainty quantification, but this is biased to conceptions of how uncertainty is perceived in 1 or 2 dimensions (and these models do not scale well). As the correlation metric shows, uncertainty quantification can take many forms with similar quantification of the underlying error as shown in figure \ref{fig:MMT_Op_Projs}. Any one model is not superior to another through direct comparison of predictive variance. Instead, models must be directly assessed against their own inherent deficiencies. If a model consistently reflects correlation of error with predictive variance, then it honestly informs the user of its deficiencies.

\textbf{Shallow ensembles of DNNs and DNOs provide encouraging predictive variance structure and distribution with respect to the true error.} Here we employ an ensemble approach consisting of only 10 members, yet the test accuracy and structure of the predictive variance are often superior in MSE and correlation metric than GPs. This is especially true for problems of higher complexity (e.g. NLS and figure \ref{fig:MMT_Oscil_Projs}~$b)$) and greater dimensionality. Considering computational cost for training $N$ members is a critical disadvantage for ensemble NNs, these results are promising for those aiming to apply ensemble NNs or DNOs to BED or BO.


\appendix

\section{Applications} \label{app:Applications}
\subsection{Stochastic Oscillator} 
Investigated previously by \cite{mohamad2018sequential} and \cite{blanchard2020output}, the stochastic oscillator is described as
\begin{equation}
    \frac{d^2s}{dt^2} + \delta \frac{ds}{dt} + F(s) = u(t),
\end{equation}
where $s(t) \in \mathbb{R}$ is the state variable, $u(t)$ is a stationary stochastic process with correlation function $\sigma_{u}^{2} \exp [-\tau^{2} / (2 \ell_{u}^{2})]$, and $F$ is a nonlinear restoring force defined by
\begin{equation}
    F(u)=\left\{\begin{array}{ll}
    \alpha s, & \text { for } 0 \leq|s| \leq s_{1} \\
    \alpha s_{1}, & \text { for } s_{1} \leq|s| \leq s_{2} \\
    \alpha s_{1}+\beta\left(s-s_{2}\right)^{3}, & \text { for } s_{2} \leq|s|
    \end{array}\right.
\end{equation}
The remaining parameters take the values, $\delta=1.5, \alpha=1, \beta=0.1, s_{1}=0.5, s_{2}=1.5, \sigma_{\xi}^{2}=0.1, \ell_{\xi}=4$, and $T=25$. The specific output of interest, shown in figure \ref{fig:2D_Oscil} is the mean value of $u(t)$ over the interval $[0, T]$ :
\begin{equation}
    f(\boldsymbol{\theta})=\frac{1}{T} \int_{0}^{T} s(t ;\boldsymbol{\theta}) \mathrm{d} t.
\end{equation}

\subsection{Nonlinear Schr\"{o}dinger Equation}
We implement a version of the nonlinear Schr\"{o}dinger (NLS) equation, supplemented with a dissipation term for stability proposed by Majda, McLaughlin, and Tabak \citep{majda1997one} for studying $1 \mathrm{D}$ wave turbulence. It is a one-dimensional, dispersive nonlinear prototype model with intermittent events described by
\begin{equation}
    i u_{t}=\left|\partial_{x}\right|^{\alpha} u+\lambda\left|\partial_{x}\right|^{-\beta / 4}\left(\left.\left.|| \partial_{x}\right|^{-\beta / 4} u\right|^{2}\left|\partial_{x}\right|^{-\beta / 4} u\right)+i D u,
    \label{eqn:MMT}
\end{equation}
where $u$ is a complex scalar, exponents $\alpha$ and $\beta$ are chosen model parameters, and $D$ is a selective Laplacian. See \citep{majda1997one,cai1999spectral} for details on the rich dynamics and \citep{zakharov2001wave,zakharov2004one,pushkarev2013quasibreathers} for its application in understanding extreme rogue waves. We refer the reader to \cite{cousins2014quantification} for details in computing this version of the NLS, but note the chosen parameters for that description here: $\alpha = 1/2$, $\beta = 0$, $\lambda = -0.5$, $k^* = 20$, $f(k) = 0$, $dt = 0.001$, and a grid that is periodic between 0-1 with $N_x = 512$ grid points. 

To propose a stochastic and complex initial condition, $u(x,t=0)$, we use the complex-valued kernel
\begin{equation}
 k(x,x^\prime)   = \sigma_{u}^2 e^{i(x-x^\prime)} e^{-\frac{(x-x^\prime)^2}{\ell_u}},
\end{equation}
with $\sigma_u^2 = 1$ and $\ell_u = 0.35$. The objective function we define for the NNs is 
\begin{equation}
    f(\boldsymbol{\theta}) = || Re(u(x,t=T;\boldsymbol{\theta}))||_{\infty},
\end{equation}
where $T=20$, while the objective function for DNOs is simply $Re(u(x,t)$.

\section{Metrics} \label{app:Metrics}
All typically used metrics are discussed here where $y_n$ refer to test samples.

\textbf{Average marginal log-likelihood (LogLL)}: Maximize
\begin{equation}
    \frac{1}{N} \sum_{n=1}^{N} \log(p(y_{n}|\theta_n))
\end{equation}

\textbf{Normalized Mean Squared-Error  (MSE)}: Minimize
\begin{equation}
    \frac{1}{N} \sum_{n=1}^{N} (\mu(\theta_n) - y_{n})^2
\end{equation}

\textbf{Prediction Interval Coverage Probability (PICP)}: Values close to $0.95$
\begin{equation}
\frac{1}{N} \sum_{n=1}^{N} \mathds{1}_{y_{n} \leq \tilde{y}_{n}^{97.5}} \cdot \mathds{1}_{y_{n} \geq \tilde{y}_{n}^{2.5}}
\end{equation}

\textbf{Mean Prediction Interval Width (MPIW)}: Minimize
\begin{equation}
\frac{1}{N} \sum_{n=1}^{N}\left(\tilde{y}_{n}^{97.5}-\tilde{y}_{n}^{2.5}\right)
\end{equation}
where $97.5$ and $2.5$ refer to percentiles of the posterior distribution, $\tilde{y}_{n}$.

\bibliography{references}

\end{document}